# When frequency domain classification is superior to spatial

**Yanir Harel[a,1], Shalom Rosner[a,1], Ronit D. Gross[a], Tal Halevi[a], Shilo Glam[a], Ella Koresh[a] and Ido Kanter[a,b,*]**

[a]Department of Physics, Bar-Ilan University, Ramat-Gan, 52900, Israel.

[b] Gonda Interdisciplinary Brain Research Center, Bar-Ilan University, Ramat-Gan, 52900, Israel.

*Corresponding author at: Department of Physics, Bar-Ilan University, Ramat-Gan, 52900, Israel. E-mail address: ido.kanter@biu.ac.il (I. Kanter).

[1]These authors equally contributed to this work

**Abstract**

Modern image classification models largely rely on local characteristics and feature extraction for their predictions. This focused approach works well in the spatial domain, but reportedly falls short in the frequency domain, even though both representations contain the same information. This study demonstrates that the significant reduction in accuracy in the frequency domain stems from the lack of local features. Locally swapping data in the frequency domain significantly decreases its inverse spatial domain accuracy, even below that of the frequency domain; however, the frequency domain accuracy remains relatively constant. Moreover, the use of datasets containing global features in the spatial domain yields higher accuracy in the frequency domain. Using these results, a hybrid model is constructed, combining one model trained only in the frequency domain and another trained only in the spatial domain to achieve higher classification accuracy. The proposed visual geometry group (VGG) and vision transformer (ViT) architectures are trained on the ImageNet-100, WikiArt, DTD, FMD, PSVRT, and KTH-TIPS2-b datasets and evaluated.

## 1. Introduction

Fourier transform (FT) serves as a foundational alternative method for encoding information that translates standard data into the frequency domain. Its discrete version, discrete FT (DFT), is a mathematical mechanism that translates raw, complex data from temporal or spatial domains into distinct frequency components. This alternative approach to encoding information unlocks highly efficient methods for analyzing, compressing, and manipulating data. Consequently, this representation has been utilized in various fields, including physics[1,2], biology[3], cybersecurity[4], speech processing[5], face detection[6], optics, and engineering[7].

For instance, DFT is often applied to images, moving from the spatial domain to the frequency domain. From an information perspective, both representations essentially encode the same information and exhibit the same number of components because the exact original data can be recovered using inverse fast Fourier transform (IFFT). Theoretically, this implies that training a neural network on a dataset of transformed images should yield the same classification accuracy. However, previously reported results indicate that the achievable spatial domain accuracy is significantly higher than that in the DFT representation[8–10].

In view of this fundamental, significant performance gap, researchers use the frequency domain for its computational efficiency, compression, and improved runtime, instead of using it to extract useful information. For example, some methods extract Discrete Cosine Transform (DCT) coefficients directly from compressed JPEGs, or apply DCTs strictly to local $8 \times 8$ blocks, allowing models to prune high-frequency noise aggressively while preserving macro-spatial layouts[10,11]. Although previous studies have leveraged the frequency domain for anti-aliased downsampling (Spectral Pooling)[12], and deployed specialized architectures, such as FFC[13] and GFNet[14], to utilize the frequency domain for mathematical shortcuts while performing the original spatial localized convolution in the frequency space, they have not addressed the accuracy gap between the two domains.

This observation raises the following questions:

- Is there truly a performance gap in image classification tasks that is reflected in the difference between the achievable accuracies in the spatial and frequency domains?

- If this gap is unavoidable, what is its underlying reason?
- Based on the identified underlying reason, are there datasets on which the accuracy in the frequency domain is higher than that in the spatial domain?

This paper explores these questions using Convolutional Neural Networks (CNNs)[13,14] and ViTs[15,16] trained on various classification image datasets. In particular, VGG[19] and convolutional compact transformer (CCT) architectures[20] are trained on ImageNet[5,6], WikiArt[23], DTD[24], FMD[25], PSVRT[26], and KTH-TIPS2-b[27] classification image datasets in both the spatial and frequency domains.

The leading hypothesis is that the reduction in accuracy occurs because standard models are explicitly built to capture local spatial features that the frequency domain inherently lacks. Structured, multi-step analysis is performed in this paper. First, using localized transformation of small patches to the frequency domain, recovery of spatial domain accuracy is demonstrated to be viable. Moreover, accuracy is observed to decrease as the size of the patches is increased, indicating a monotonic interpolation between the frequency and spatial accuracies. Next, local data-swapping techniques are applied in the frequency domain, inducing a steep reduction in accuracy in the spatial domain, while that in the frequency domain remained relatively constant. Further, specific datasets that naturally excel in the frequency domain owing to structures dominated by global spatial features and lacking local features are identified. Finally, a hybrid model is proposed that leverages these insights, extracts supplementary global frequency information, and tightly fuses it with local spatial data to improve overall model accuracy.

## 2. Significant gap between frequency and spatial domain accuracies

The difference between the spatial and frequency domain accuracies is demonstrated using ImageNet-100 dataset classification. The frequency domain is calculated using fast Fourier transform (FFT) and represented in terms of its real and imaginary forms. When CNNs, such as VGG-14 (13 convolution layers followed by a single fully connected layer)[28] are trained on the dataset, the gap in test accuracy is $\sim$0.13 for 10 classification labels and increases to $\sim 0.3$ for 100 labels (Fig. 1). This massive reduction in test accuracy is highly counterintuitive, given that both representations contain mathematically identical information as evidenced by the recovery of the original image in the spatial domain via IFFT. This substantial accuracy gap is also experimentally confirmed for a ViT architecture of CCT-7/7x2 (Fig. 2). Further, the results confirm that this accuracy gap persists even for models across varying numbers of labels selected from the dataset (Fig. 1-2).

**ImageNet-100**

| No. of labels | Accuracy | |
|---|---|---|
| | Spatial domain | Frequency domain |
| 10 | 0.934 | 0.802 |
| 30 | 0.875 | 0.629 |
| 50 | 0.850 | 0.604 |
| 70 | 0.837 | 0.569 |
| 100 | 0.794 | 0.506 |

**Fig. 1.** Test accuracies of VGG-14 on the ImageNet dataset in the spatial and the frequency domains with respect to varying numbers of labels.

| ImageNet-100 | | |
|---|---|---|
| No. of labels | Accuracy | |
| | Spatial domain | Frequency domain |
| 10 | 0.894 | 0.766 |
| 30 | 0.827 | 0.629 |
| 100 | 0.711 | 0.467 |

**Fig. 2.** Test accuracies of CCT-7x2 on the ImageNet dataset in the spatial and the frequency domains with respect to varying numbers of labels.

Reduced test accuracy in the frequency domain has been reported in previous studies [8,9]. Although the accuracies in the frequency domain reported in Figures 1–2 are slightly higher compared to those in previous studies, the gap is still significant. Thus, the question that arises is whether this gap in test accuracy stems from fundamental differences between their representations or from the distinct learning strategies required in each domain.

## 3. Recovered accuracy by combining the two domains

The proposed hypothesis is that the gap in test accuracy originates from the fact that the frequency domain mostly encodes global features of images, whereas the spatial domain encodes local features. To test this hypothesis, the data is processed by combining both domains, allowing the frequency representation to also capture some local features. Specifically, the image is divided into small partially overlapping patches, where each patch is represented in the frequency domain (Fig. 3a). The results indicate that the accuracy in the spatial domain can be recovered using the small patches of frequency domain data. For instance, dividing a spatial picture into $8 \times 8$ overlapping patches is observed to recover the test accuracy to that of the image in the spatial domain. Specifically, each $224 \times 224$ image is split into $8 \times 8$ patches with a stride of 2 pixels, resulting in an area larger by a factor of $\sim 16$ $(872 \times 872)$. Next, each $8 \times 8$ patch is converted to its frequency domain and downsized using $4 \times 4$

MaxPooling to approximately the same size as the original image (Fig. 3b). The size of the final image $o^*$ is calculated as follows:

$$o^* = \left(\frac{o - p}{s} + 1\right)\frac{p}{m} \tag{1}$$

where $o$ denotes the original image size, $p$ denotes the patch-window size, $s$ denotes the stride, and $m$ denotes the MaxPooling window size.

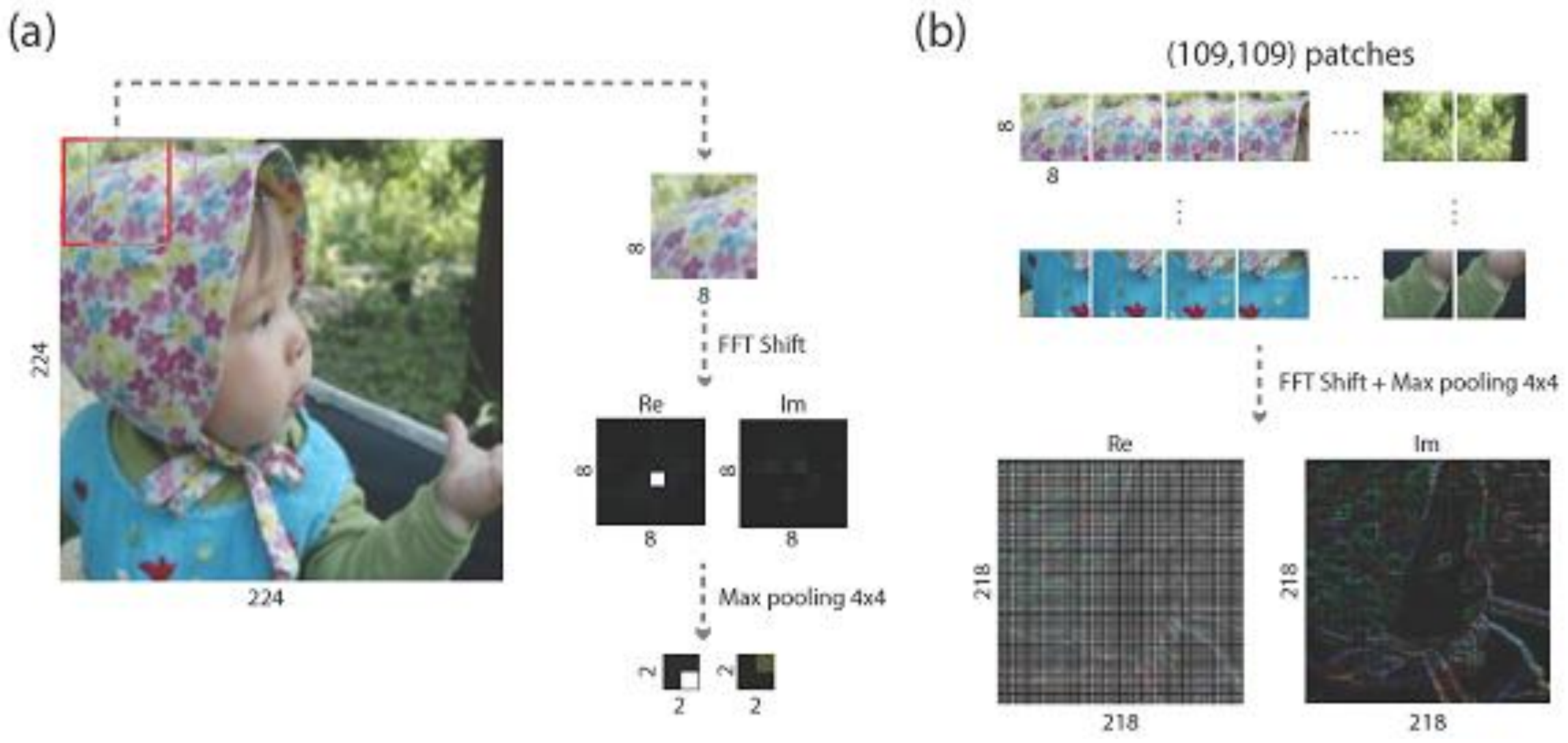

**Fig. 3** (a) Process scheme combining spatial and frequency domains. A $224 \times 224$ image is split into patches using $p = 8$ and $s = 2$, creating distinct maps for the real and imaginary components, following by downscaling with $m = 4$ (Eq. 1) (b) Construction scheme of the full $218 \times 218$ new image by applying IFFT to the computed patches.

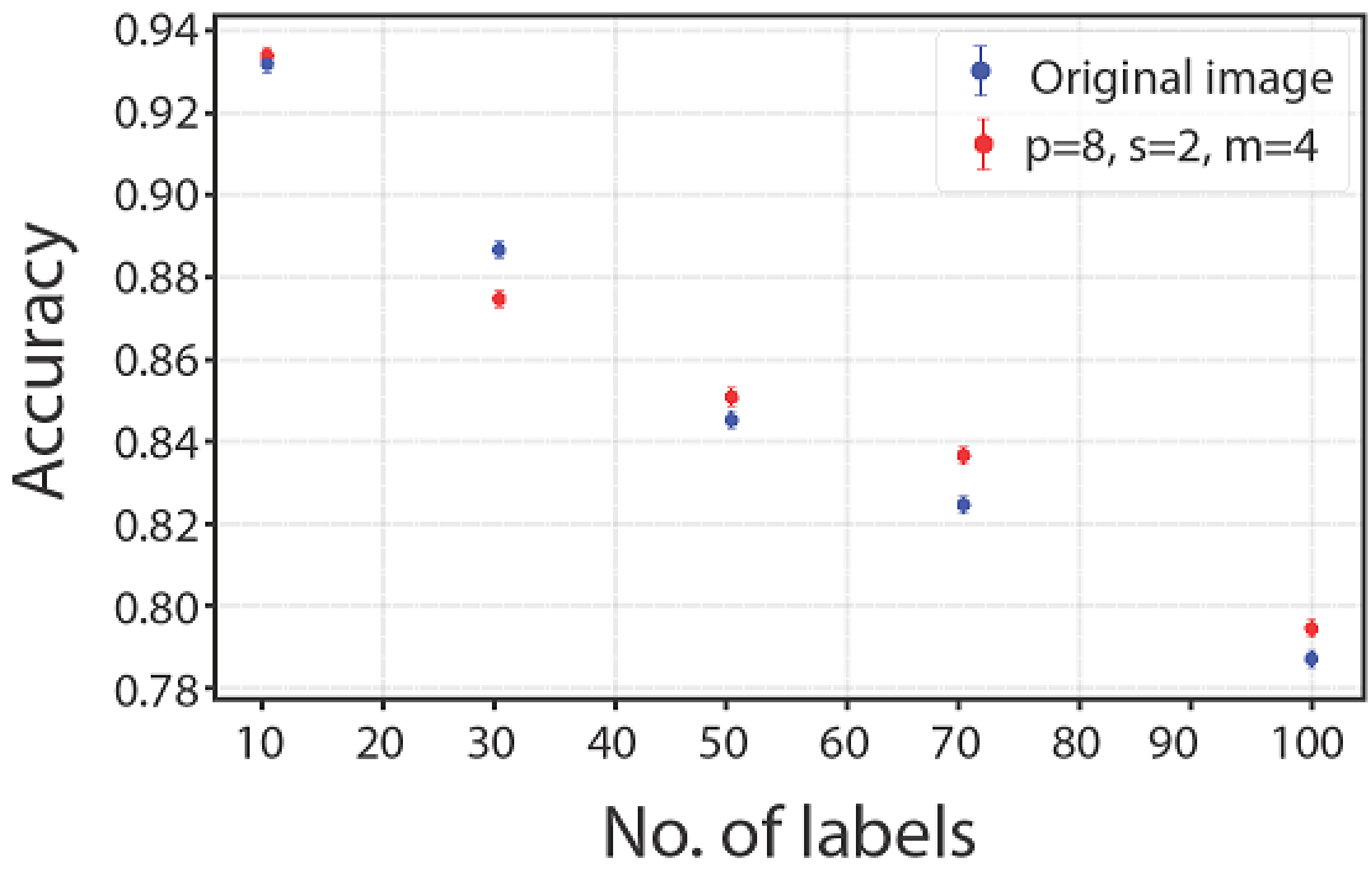

**Fig. 4.** Comparison of test accuracies using VGG-14 along with their error bars for $p = 8$, $s = 2$, and $m = 4$ FFT patches and the original images with respect to varying numbers of labels selected from the Imagenet-100 dataset. The size of the input image created by the patches is $o^* = 218$ (Eq. 1), which is relatively close to that of the original image. i.e., $o = 224$.

When $p = 8$, $s = 2$, and $m = 4$, the test accuracies corresponding to the combined spatial and frequency domains, are nearly identical and independent of the number of trained labels (Fig. 4). However, the accuracy decreases monotonically with increasing patch size until it reaches the accuracy of the image in the frequency domain (Fig. 5). These results clearly indicate that combining frequency and spatial representations on a localized scale can achieve comparable results as in the spatial domain. However, the significant gap between the accuracies in the two domains when the entire images are examined ($p = 224$) remains unexplained.

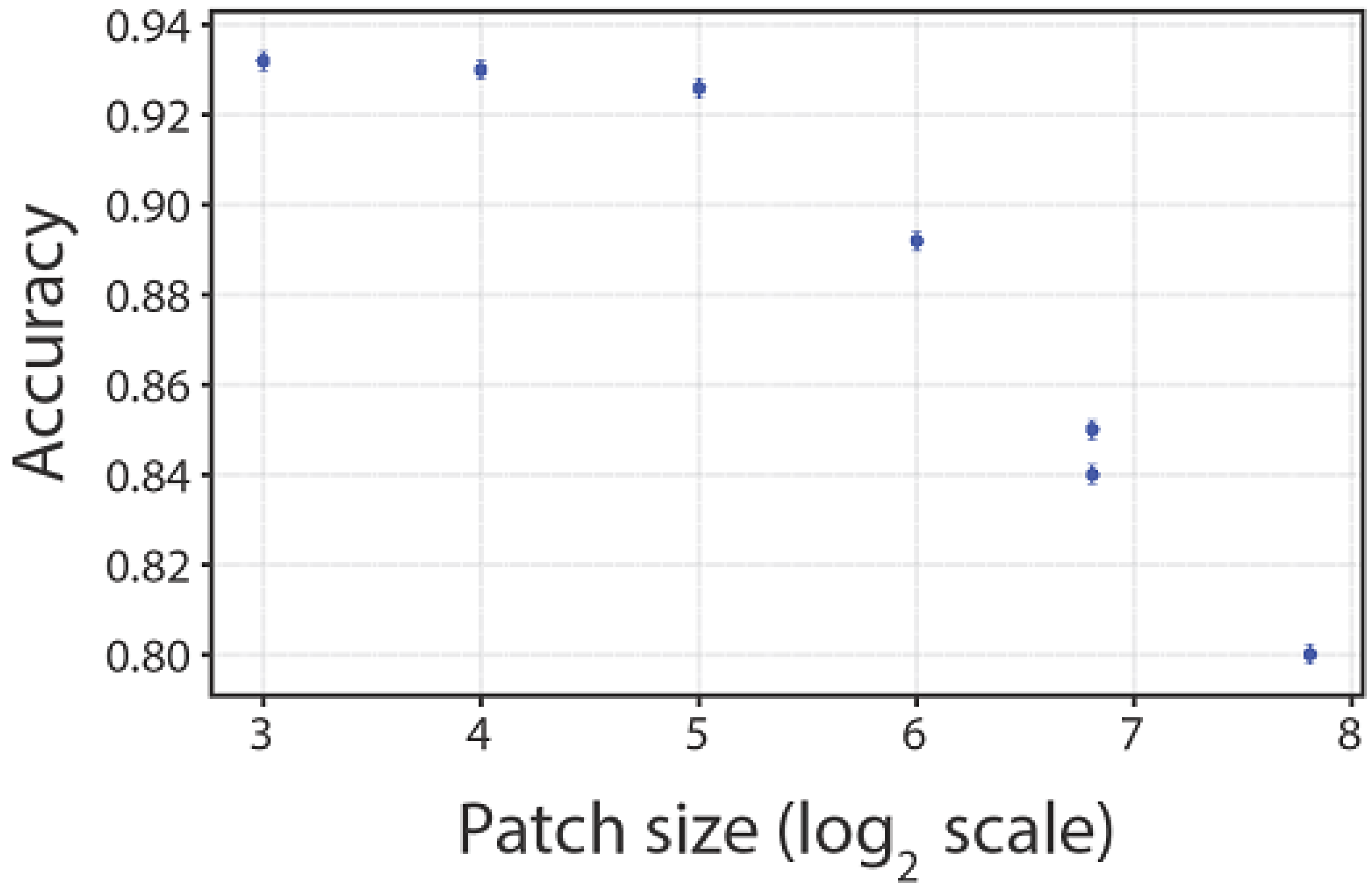

**Fig. 5.** Test accuracies and corresponding error bars using VGG-14 trained on ImageNet dataset with 10 labels as a function of the patch size and fixed $s = 2$. The x-axis is presented in $log_2$, since for small patch sizes it is doubled for each subsequent measurement. When $p = 112$, $m = 28$, and $m = 56$ yield nearly identical accuracies, albeit with images sizes of $o^* = 112$ and $o^* = 56$, respectively. This indicates that the reduction in accuracy is mainly attributed to the patch size.

## 4. Underlying reason for the accuracy gap

To validate the hypothesis that the accuracy gap stems from the lack of localized features in the frequency domain compared to the spatial domain, the images in the frequency domain are locally scrambled within each of the non-overlapping patches (a fixed scrambling scheme is applied to all images). As the transformed matrix is self-conjugated, each selected frequency coefficient and its corresponding conjugate coefficient move simultaneously, thereby preserving the conjugate symmetry of the Fourier representation. The self-conjugate coefficients, including the Nyquist frequencies, remain fixed and are not subjected to mixing. Next, the accuracy is measured on the locally scrambled dataset in the frequency domain and compared with the corresponding accuracy on the spatial dataset following IFFT, in the complete absence of local features (Fig. 6). The results indicate that the accuracy in the frequency domain is affected only slightly, whereas that in the spatial domain decreases significantly. Moreover, the accuracy in the frequency domain is higher than that in the spatial domain (Fig. 7). This trend is confirmed for varying numbers of labels for both the CCT and VGG trained architectures. Similar results are also obtained when scrambling is not confined within individual patches, e.g., when local pairs of real and imaginary components are exchanged in the frequency domain (See Appendix).

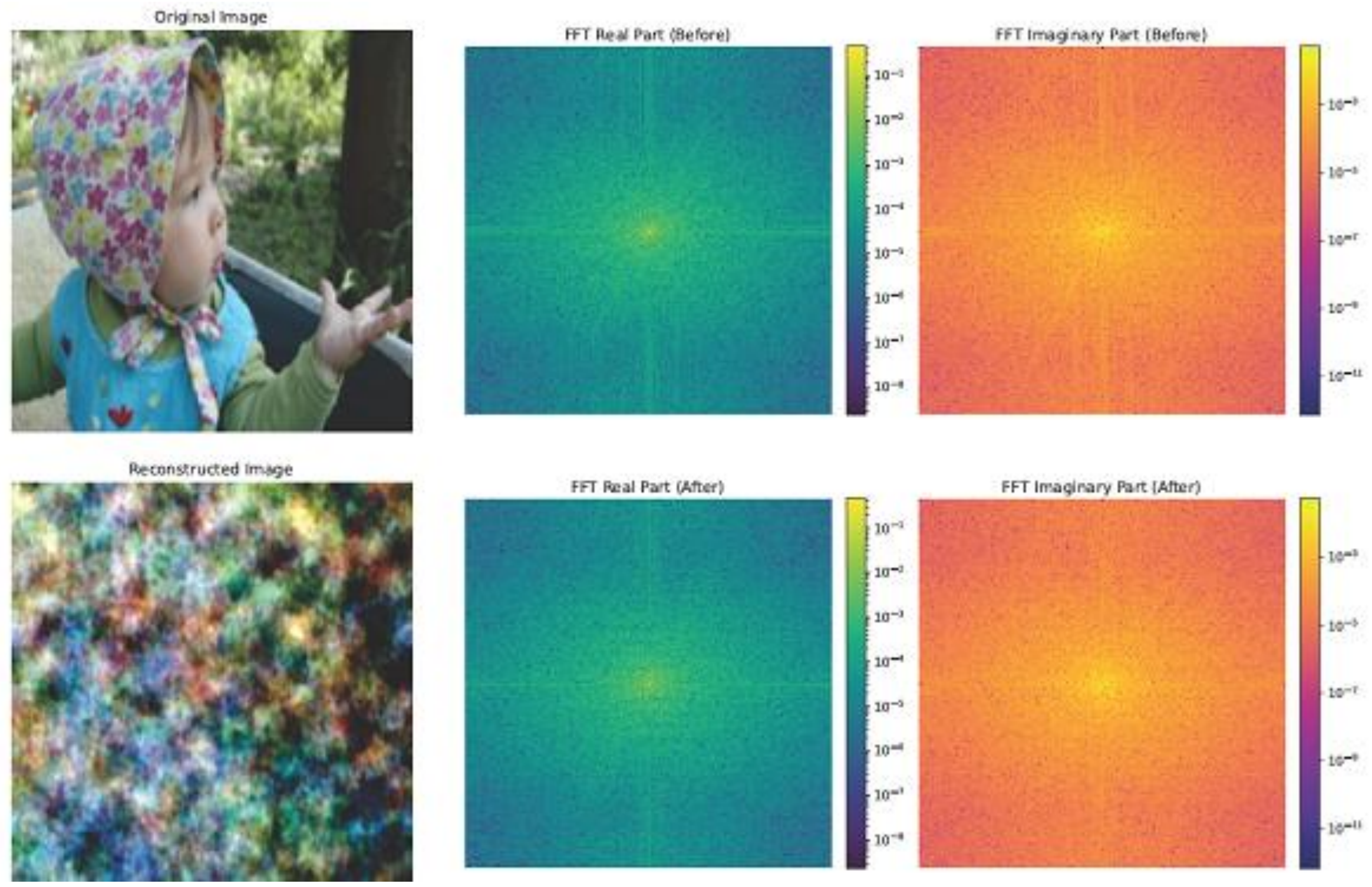

**Fig. 6.** Example of an original ImageNet image (left-top) and the IFFT of the scrambled spatial image after randomly scrambling of each patch in the frequency domain, with p = 8 and s = 8 (Eq. 1) (bottom-left). Average real (second column) and imaginary (third column) components using FFT presented in its shifted form (FFT-Shift).

| Image setting | | | | Dataset | | |
|---|---|---|---|---|---|---|
| FFT | mixing | Patch size | IFFT | ImageNet-10 | ImageNet-30 | ImageNet-100 |
| X | X | X | X | 0.942 | 0.875 | 0.823 |
| V | X | X | X | 0.742 | 0.614 | 0.447 |
| V | V | 5x5 | V | 0.804 | 0.665 | 0.545 |
| V | V | 5x5 | X | 0.766 | 0.626 | 0.477 |
| V | V | 8x8 | V | 0.732 | 0.584 | 0.433 |
| V | V | 8x8 | X | 0.756 | 0.622 | 0.470 |

**Fig. 7.** Accuracies using VGG-16 on it spatial version (first row) and on its FFT version (second row) corresponding to varying number of labels selected from the spatial ImageNet dataset. Accuracies obtained for mixing within each $5 \times 5$ patches in the frequency and the spatial domains (third-fourth rows) and, similarly, for mixing $8 \times 8$ patches (fifth-sixth rows).

These results clearly demonstrate a significant decrease in the accuracy measured in the spatial domain when local features are significantly distorted by local scrambling in the frequency domain. Conversely, the accuracy achieved in the scrambled frequency domain is only mildly affected. The accuracy achieved in the spatial domain is highly dependent on local features.

## 5. Superior frequency domain accuracy on datasets with global features

To further demonstrate the leading claim of the importance of local features in achieving high accuracy in the spatial domain, datasets containing primarily global features are examined. The results indicate that the lack of dominant localized features in these datasets reduce the accuracy in the spatial domain even below that of the frequency domain (Fig. 8). The first example is the texture classification dataset, KTH-TIPS2-b, which contains images of 11 different materials (Fig. 9). The second dataset is PSVRT, on which a classification task is performed to determine whether or not two objects in a single image are the same. This task requires global feature extraction because the objects are not always located near each other in the image.

| Dataset | Test accuracy | |
|---|---|---|
| | Spatial domain | Frequency domain |
| KTH-TIPS2-b | 0.612 | 0.696 |
| PSVRT | 0.932 | 1.000 |

**Fig. 8.** Test accuracies using VGG-14 architecture trained on KTH-TIPS2-b and PSVRT datasets in the spatial and frequency domains.

These results verify the hypothesis because the frequency domain accuracy exceeds that in the spatial domain. This trend indicates that when the task is governed primarily by global characteristics, the accuracy gap between the spatial and frequency representations decreases, with the balance even shifting in favor of the frequency domain.

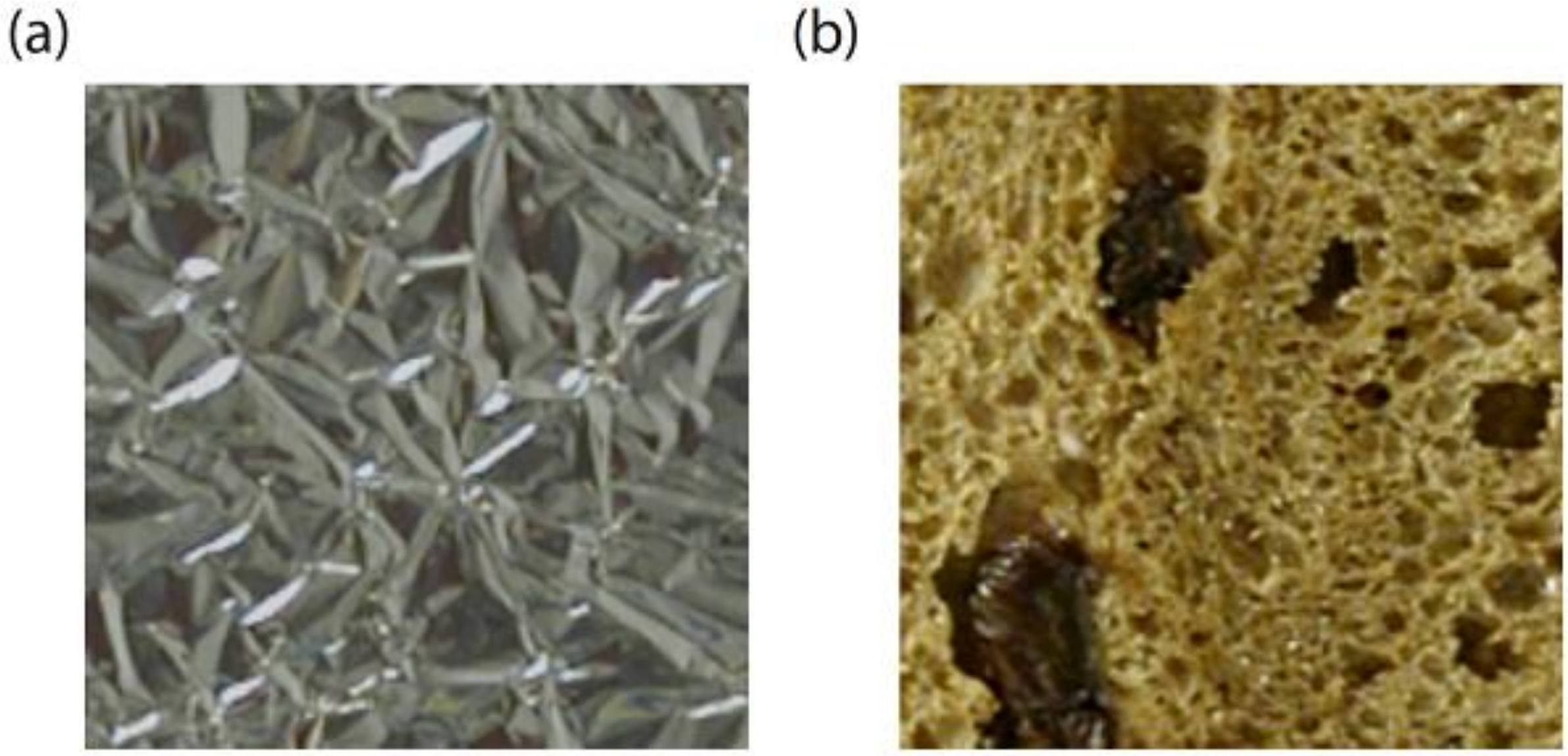

**Fig 9.** (a) Sample image from the class "aluminium_foil" from the KTH-TIPS2-b dataset. (b) Sample image from the class "brown_bread" from the same dataset. Images from this dataset classify texture and contain global characteristics.

## 6. Superior accuracy on datasets combining classes with local and global features

Most datasets usually comprise a variety of classes, with some having more localized features and others more global features. One challenge is to determine training schemes that improve the accuracy in such hybrid datasets. An example of such a hybrid dataset is WikiArt, where some styles are dominated by global features and most by local features (Fig. 10). The VGG-14 architecture is trained on the dataset after mitigating its class imbalance by setting the sample count for each label to be equal to that of the minority class via under-sampling. This subset of the dataset is trained once each in the spatial and frequency domains. The confusion matrix indicates four labels with enhanced accuracy in the frequency domain (see Appendix), which are characterized by dominant global features.

To leverage the different characteristics governing the labels, the following hybrid network is implemented. Two identical architectures (VGG-14 in the following example) are trained independently—one using the dataset in the frequency domain, and the other using that in the spatial domain. The dataset consists of four labels that perform better in the frequency domain, along with four others with enhanced accuracy in the spatial domain (see Appendix). Only four are selected to balance the effect of each domain. The predicted output labels are selected based on confidence of the two independently trained networks. The results indicate improved accuracy compared to the network trained only on the spatial domain dataset.

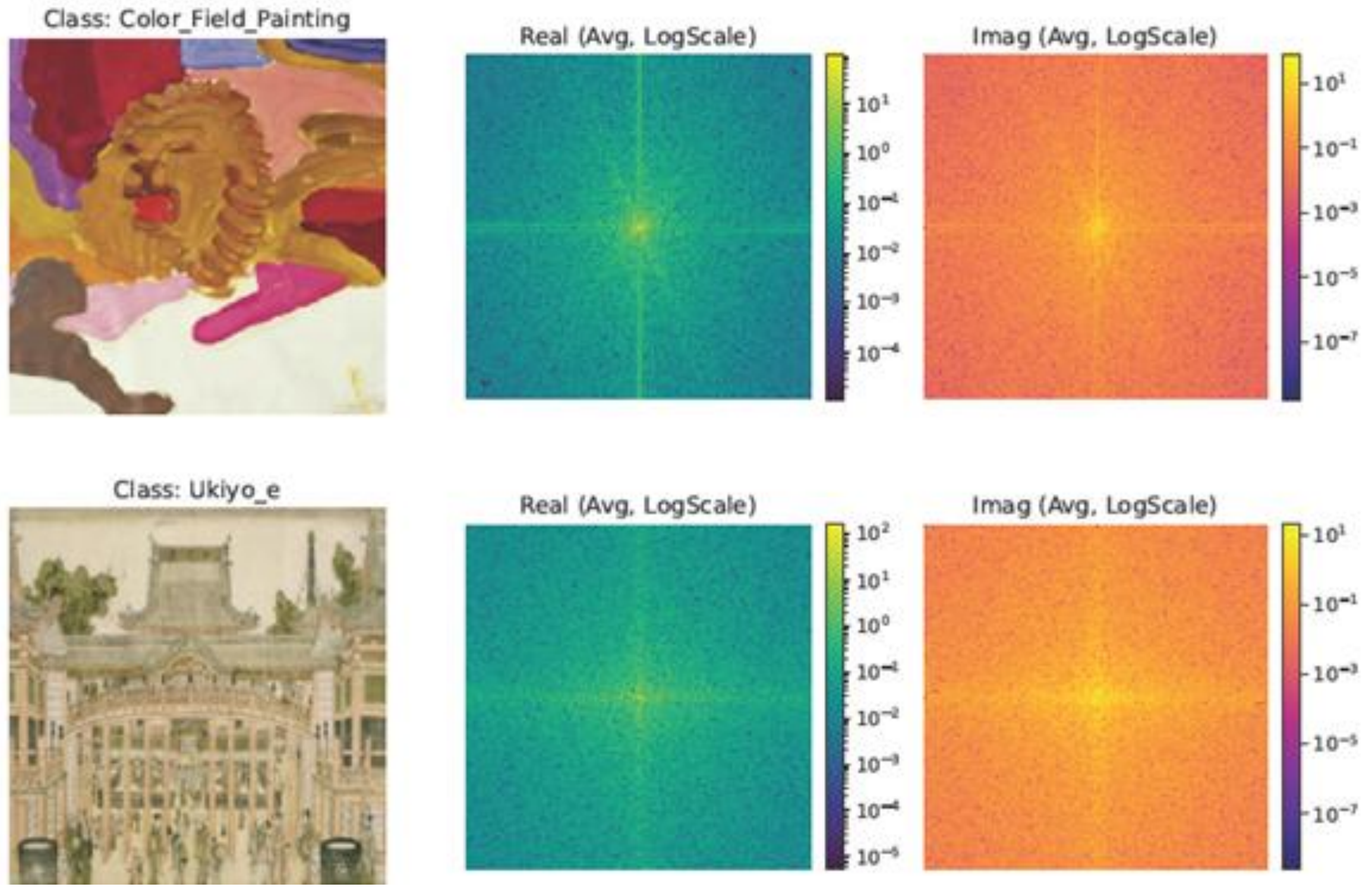

**Fig. 10.** (a) The original image from the 'Color_Field_Painting' class in the WikiArt dataset (left), which mainly comprises global features, its real value in the frequency domain averaged over the three-color channels in log-scale (middle), and its averaged imaginary value in log scale (right). (b) Same as (a) but for image from the class 'Ukyo_e'.

To validate accuracy improvement using the hybrid network, results are derived on other datasets, such as DTD and FMD. While both DTD and FMD contain texture images, some of their classes contain zoomed-out objects with some localized features, improving their accuracy in the spatial domain. Validation on these datasets is performed using the same pipeline as that on the WikiArt dataset. First, a VGG network is trained on these datasets and the confusion matrix is examined. Next, the labels that perform better in the frequency domain are selected (see Appendix for selected labels). Finally, the hybrid network is trained, and accuracy is observed to improve compared to the network trained only on spatial domain data (Fig. 11).

| Dataset | Test accuracy | | |
|---|---|---|---|
| | Spatial domain | Frequency domain | Hybrid model |
| WikiArt | 0.638 | 0.625 | 0.679 |
| DTD | 0.673 | 0.667 | 0.708 |
| FMD | 0.417 | 0.353 | 0.447 |

**Fig. 11.** Accuracies using VGG-14 architectures trained in the spatial and frequency domains, and that of the hybrid model trained on both domains independently, evaluated on subsets of the three datasets.

## 7. Discussion

In the recent years, significant progress has been made in applications of machine learning in image processing. However, the frequency domain is typically not used in this context, despite its ubiquity in other fields [1,2],[3],[4],[5], [6],[7]. This is primarily because of the significant reduction in the achievable accuracy in the frequency domain compared to the spatial one. Determining whether this gap is unavoidable and, if so, understanding its underlying reason is a significant step in utilizing this representation comprehensively.

This paper demonstrates that although spatial and frequency representations contain the same information from an informatics perspective, they are fundamentally expressed in different ways. In the frequency domain the information is organized based on global features stemming from the fact that frequency is a global measure. In contrast, the information in the spatial domain is sensitive to local features. This informatics perspective is first supported by the result that the achievable accuracy using small patches in the frequency domain is the same as that of the spatial domain (Fig. 7). Here, local features confined to small patches are captured well in both the spatial and frequency domains (Fig. 3). Upon enhancing the weight of global features by increasing the patch size, a monotonic increase is observed in the gap between the accuracy in the spatial and frequency domains (Fig. 5). Moreover, locally swapping data in the frequency domain disrupts the local features significantly. This type of local perturbation results in a significant reduction in spatial domain accuracy; however, the frequency domain accuracy remains relatively constant.

In further corroboration of the informatics insight, images corresponding to labels mostly governed by local features are observed to yield superior accuracy in the spatial domain, whereas images belonging to those mainly governed by global features yield enhanced accuracy in the frequency domain. Based on this insight, the proposed hybrid-trained model is designed to leverage the benefits of both representations, improving the accuracy compared to models trained on a single representation (Fig. 11).

Ultimately, these findings suggest a promising theoretical direction for the development of energy-based models. Similar to the use of momentum space and renormalization group methods in statistical mechanics to separate global structures from local noise, translating dreaming Hopfield networks [27,28,29] and bidirectional associative memories[30,31] to the frequency domain could provide a framework for improving pattern recall.

Finally, this paper demonstrates that some information does not overlap when models are trained on frequency and spatial domain data, indicating that some data crucial for the classification task are not fully extracted from the image. This calls for a more sophisticated model architecture capable of leveraging both global and local characteristics. The model should apply the underlying idea of the hybrid model proposed in this paper in a broader manner, fully utilizing the data by configuring the model's structure in a way that also extracts information from the frequency representation, just as CNN models are explicitly structured to extract local features.

Recent studies have indicated that internal learning dynamics drive network components, such as filters, heads, and nodes, to specialize in distinct labels[29,30,31,32,33]. Consequently, another open research question is whether these components can be guided to specialize in labels governed by local and global features separately to utilize both representations during model training. Further, different attention heads within an attention block are known to focus on different linguistic characteristics[39]. Thus, performing the same analysis on CCT models trained on the frequency-domain data might help identify more informative frequencies for capturing global features of images.

The results presented in this paper are based on two benchmark models that use only image classification tasks. Thus, further validation using other types of data, such as audio and other signal processing tasks, is required.

**Acknowledgements**

The work is supported by Ministry of Innovation, Science and Technology [grant number 0008545].

## Appendix

### *1. Datasets and preprocessing*

The datasets used in this study are ImageNet-100[10,11], WikiArt[23], DTD[24], FMD[25], PSVRT[26] and KTH-TIPS2-b[27]. The subset of labels from the ImageNet-100 dataset were selected by the order in which they appear in the dataset. Each dataset before training was preprocessed on the following pipeline, first applies spatial data augmentation consisting of random resized cropping to $224 \times 224$ pixels[40] and random horizontal flipping[41]. Following tensor conversion, the images undergo global dataset normalization[42]. These steps were applied to all the training data, when the data was transformed to it frequency domain the Fast Fourier transformation (FFT)[43] was used to Reduce the complexity of DFT. When presenting the data in the figures (Fig. 6,10) above we used the FFT-shift where the low frequencies are placed in the center instead of the corners.

### *Architectures*

Three different architectures were examined, VGG-16[44], VGG-14[19] and CCT-7/7x2[20,45]. The use of VGG14 was mainly to speed up the training process, however, there seems to be no noticeable difference in performance of both VGG models. Both VGG architectures where trained using the D=16 (initial number of filters in the conv-3 layer). When The models where fed the frequency representation there were 6 input channels, R/G/B for the real component and R/G/B for the imaginary component.

### *2. Optimization*

The CrossEntropyLoss[46] function was selected for the classification task and minimized using the stochastic gradient descent algorithm[17,18] The maximal accuracy was determined by searching through the hyper-parameters (Fig. 12). The L2 regularization method[49] was applied along with the use of the classical momentum[50]

### *3. Hyper-parameters*

The hyper-parameters $\eta$ (learning rate), $\alpha$ (L2 regularization) and $\beta$ (momentum) were optimized for offline learning, using a mini-batch size of $128$ inputs. The learning-rate decay schedule was also optimized. For all simulations, a cosine annealing scheduler was applied [41,42,43], with zero warm-up steps and the total number of training steps set equal to the number of epochs. This schedule gradually decays the learning rate from its initial value to zero following a cosine curve throughout training, which helps stabilize convergence. The models in all the simulations were trained for 150 epochs or 300 epochs (when it seems the train accuracy didn't fully converge).

| Hyper-parameters | | | | |
|---|---|---|---|---|
| | η | α | β | epochs |
| *Figure 1* | 0.03 | 4e-5 | 0.9 | 150 |
| *Figure 2* | 0.03 | 4e-5 | 0.9 | 300 |
| *Figure 4-5* | 0.03 | 4e-5 | 0.9 | 150 |
| *Figure 7* | 0.03 | 4e-5 | 0.9 | 300 |
| *Figure 8* | 0.03 | 4e-5 | 0.9 | 150 |
| *Figure 11* | 0.03 | 4e-5 | 0.9 | 150 |
| *Figure 13* | 0.03 | 4e-5 | 0.9 | 300 |
| *Figure 14* | 0.016 | 4e-5 | 0.9 | 300 |

**Fig 12.** Hyperparameters used for training each model, corresponding to the results presented in each Figure.

### *4. Mixing result on CCT model*

Figure 13 below presents the identical trends observed in the VGG model using the CCT model, demonstrating that the decrease in accuracy is much smaller in the frequency domain than that in in the spatial domain (Fig. 7).

| Image setting | | | | Dataset | | |
|---|---|---|---|---|---|---|
| FFT | mixing | Patch size | IFFT | ImageNet-10 | ImageNet-30 | ImageNet-100 |
| X | X | X | X | 0.894 | 0.827 | 0.711 |
| V | X | X | X | 0.766 | 0.629 | 0.467 |
| V | V | 5x5 | V | 0.670 | 0.531 | 0.353 |
| V | V | 5x5 | X | 0.706 | 0.563 | 0.389 |
| V | V | 8x8 | V | 0.662 | 0.508 | 0.342 |
| V | V | 8x8 | X | 0.690 | 0.575 | 0.366 |

**Fig. 13.** Accuracies of *CCT-7/7x2* corresponding to varying number of labels selected from the spatial ImageNet dataset (first row) and on its FFT version (second row). Accuracies obtained for mixing within each $5 \times 5$ patches in the frequency and the spatial domains (third-fourth rows) and, similarly, for mixing $8 \times 8$ patches (fifth-sixth rows).

### *5. Local perturbation results*

To confirm that the reduction in accuracy is not induced by patch-based mixing, another mixing method, based on a more classic physically inspired method of perturbation, is tested on models. In this version, each pixel/frequency is swapped with a randomly selected second pixel within a 2-pixel distance of it. Once a pixel has been swapped, it cannot be swapped again—this perturbation is fixed for all samples. As the transformed matrix is self-conjugated, each selected frequency coefficient and its corresponding conjugate coefficient move simultaneously, thereby preserving the conjugate symmetry of the Fourier representation. The self-conjugate coefficients, including the Nyquist frequencies, remain fixed and are not subject to perturbations. The results exhibit the same trend (Fig. 14) as the previous method, i.e., the decrease in accuracy in the frequency domain is much smaller than that in the spatial domain (Fig. 7).

| Image setting | | | | Dataset | |
|---|---|---|---|---|---|
| FFT | mixing | IFFT | No.labels | VGG-16 | CCT |
| X | X | X | ImageNet-10 | 0.934 | 0.9 |
| V | X | X | | 0.802 | 0.766 |
| V | V | X | | 0.768 | 0.728 |
| V | V | V | | 0.724 | 0.676 |
| X | X | X | ImageNet-30 | 0.896 | 0.831 |
| V | X | X | | 0.664 | 0.628 |
| V | V | X | | 0.624 | 0.568 |
| V | V | V | | 0.595 | 0.521 |

**Fig. 14**. Accuracies using the VGG16 and *CCT-7/7x2* models corresponding to varying numbers of labels selected from the ImageNet dataset after pixel local perturbation.

## *6. Selected dataset for hybrid model*

Figure 15 depicts the selected subsets of labels used to train the hybrid model.

| Dataset | Labels | |
|---|---|---|
| | Frequency | Spatial |
| WikiArt | ”Art_Nouveau”, ”Color_Field_Painting”, ”Minimalism”, ”Rococo” | ”Baroque”, ”Expressionism”, ”Ukiyo_e”, ”Pop_Art” |
| DTD | ”marbled”, ”knitted”, ”stratified”, ”stained”, ”blotchy”, ”meshed”, ”grooved” | ”sprinkled”, ”bubbly”, ”studded”, ”cobwebbed”, ”wrinkled”, ”crystalline”, ”paisley” |
| FMD | ”plastic”, ”fabric”, ”leather” | ”water”, ”glass”, ”paper” |

**Fig. 15.** Subset of labels from WikiArt, DTD, and FMD datasets selected to test hybrid model.

### *7. Statistics*

Statistics for all results were obtained using at least five samples and the standard deviation was around 1% for all the results.

### *8. Hardware and software*

We used Google Colab Pro and its available GPUs. We used Pytorch for all the programming processes.